\pdfoutput=1

\documentclass[11pt]{article}

\usepackage{ACL2023}

\usepackage{times}
\usepackage{latexsym}

\usepackage[T1]{fontenc}

\usepackage[utf8]{inputenc}

\usepackage{microtype}

\usepackage{inconsolata}

\usepackage{enumitem}
\usepackage{graphics}
\usepackage{graphicx}
\usepackage{algorithm}
\usepackage{algorithmic}
\usepackage{subfigure}
\usepackage{amsmath}
\usepackage{amsthm}
\usepackage{booktabs}
\usepackage{multirow}
\usepackage{pifont}

\usepackage{soul}
\title{Fennec: Fine-grained Language Model Evaluation and Correction Extended through Branching and Bridging}

\author{
    Xiaobo Liang\textsuperscript{\normalfont 1} $^{\ast}$ \quad
    Haoke Zhang\textsuperscript{\normalfont 1} \thanks{\; Equal Contribution} \quad
    Helan hu\textsuperscript{\normalfont 2}\quad
    Juntao Li\textsuperscript{\normalfont 1} \thanks{\ Corresponding Author}\quad
    Jun Xu\textsuperscript{\normalfont 3}\quad
    Min Zhang\textsuperscript{\normalfont 1} \\
    \textsuperscript{1}Soochow University, \textsuperscript{2}Peking University, \textsuperscript{3}Baidu Inc. \\
    \texttt{\{xbliang3, zctang\}@stu.suda.edu.cn}, \\
    \texttt{\{huhelan\}@stu.pku.edu.cn}, \\
    \texttt{\{xujun03\}@baidu.com}, \\
    \texttt{\{ljt,minzhang\}@suda.edu.cn} \\
}

\begin{document}
\maketitle
\begin{abstract}

The rapid advancement of large language models has given rise to a plethora of applications across a myriad of real-world tasks, mainly centered on aligning with human intent. 
However, the complexities inherent in human intent necessitate a dependence on labor-intensive and time-consuming human evaluation.
To alleviate this constraint, we delve into the paradigm of employing open-source large language models as evaluators,
aligning with the prevailing trend of utilizing GPT-4.
Particularly, we present a step-by-step evaluation framework: \textbf{Fennec}, capable of \textbf{F}ine-grained \textbf{E}valuatio\textbf{N} and correctio\textbf{N} \textbf{E}xtended through bran\textbf{C}hing and bridging.
Specifically, the branching operation dissects the evaluation task into various dimensions and granularities, thereby alleviating the challenges associated with evaluation. 
Concurrently, the bridging operation amalgamates diverse training datasets, augmenting the variety of evaluation tasks.
In experimental trials, our 7B model consistently outperforms open-source larger-scale evaluation models across various widely adopted benchmarks in terms of both \textit{Agreement} and \textit{Consistency}, closely approaching the capabilities of GPT-4. 
We employ the fine-grained correction capabilities induced by the evaluation model to refine multiple model responses, and the results show that the refinement elevates the quality of responses, leading to an improvement of 1-2 points on the MT-Bench.
Our code is available at Github\footnote{\url{https://github.com/dropreg/Fennec}}.

\end{abstract}

\section{Introduction}

The inherent complexity of real-world tasks has led to a deficiency in impartial and equitable evaluations among diverse models~\cite{saunders2022self, liang2022holistic}, compelling dependence on evaluation methods that are both effective and efficient.
As an alternative to human evaluation, the use of Large Language Models (LLMs) in automatic evaluation techniques~\cite{zheng2023judging, dubois2023alpacafarm} has demonstrated improved efficiency and controllability:
(\romannumeral1) \textit{cost}: Alleviating the requirements for domain experts, thereby facilitating scale expansion.
(\romannumeral2) \textit{criteria}: Enabling the formulation of predefined evaluation criterion, reducing the ongoing cost of negotiation~\cite{markov2023holistic}.
(\romannumeral3) \textit{bias}: Mitigating model biases by refining training data, preventing the incorporation of challenging-to-rectify human biases~\cite{zheng2023judging}.
These advantages position automated evaluation techniques as an indispensable component in the future landscape of conversation tasks.
In this paper, we focus on the evaluation of the conversational ability of an LLM assistant~\cite{bai2022training, zhou2023lima}, specifically scenarios involving a human intent query and the response provided by AI.
Our goal is to evaluate how effectively these responses can better understand and fulfill the user's authentic intent and requirements.

\begin{figure}
    \centering
    \includegraphics[scale=0.27]{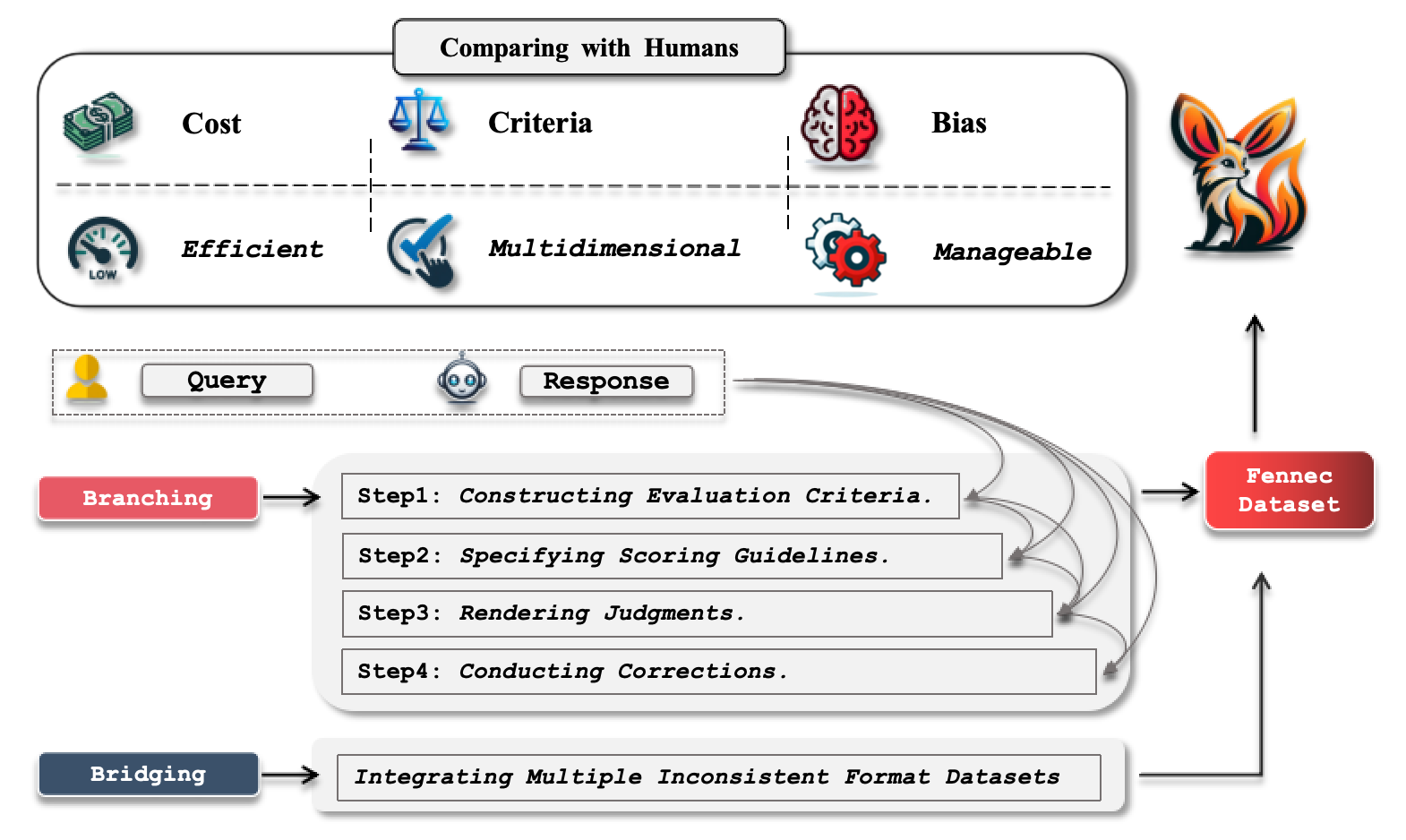}
    \caption{The illustration of Fennec showcases the construction of data for model training through branching and bridging for conversational task evaluation.}
    \label{fig:preview}
    \vspace{-0.5cm}
\end{figure}

The adoption of GPT-4~\cite{openai2023gpt4} API calls as an evaluation model or utilizing the self-instruct~\cite{wang2022self} data as a training corpus for developing new evaluation models has become a prevailing methodology.
However, such methods often overlook the inherent variations in the capabilities of the backbone models.
For example, smaller LLMs (such as the 7B variants) lack the robust commonsense and reasoning abilities presented in their larger counterparts (exceeding 100B), thereby frequently giving rise to hallucinatory outputs and even making erroneous judgments~\cite{chen2023purr}.
Our approach integrates a sequential thought methodology, like chain-of-thought~\cite{wei2022chain}, enriching the language model with contextual information to significantly augment the accuracy of its outputs. 
We employ a \textbf{\textit{branching}} operation akin to the methodologies posited in Branch-Solve-Merge~\cite{saha2023branch}, as depicted in Figure~\ref{fig:preview}. 
Concretely, the model initiates the generation of multiple \textbf{evaluation criteria} based on the user query. 
Subsequently, it formulates detailed \textbf{scoring guidelines} for these criteria. 
The evidence information generated from these intermediate processes is supplied to the evaluation model to render \textbf{judgments} for given AI responses and furnish corresponding elucidations.
The diverse comments generated can be directly fed into the evaluation model to generate \textbf{corrections} for the original responses. 
Our methodology aids the model in exploring a broader spectrum of evaluation dimensions and adapting to diverse grading granularity, thereby enhancing its adaptability across various scenarios. 
To ensure reproducibility and version control~\cite{kim2023prometheus}, we establish a new open-source training dataset and utilize it to train our evaluation model.\footnote{Our uploaded anonymous code and dataset will be released at \url{https://anonymous.com}.}

The scalability of the model is a crucial capability for addressing a variety of instructions and conversational tasks. What is especially noteworthy is the common limitation where numerous evaluation models are exclusively trained on datasets with inconsistent data formats, leading to poor scalability and even performance degradation on out-of-distribution datasets.
Therefore, a key aspect of scaling up models is the adoption of a uniform data format and the integration of multiple datasets for model training. 
We have developed a \textbf{\textit{bridging}} operation to integrate additional datasets into our training corpus. 
This involves training a reverse model using existing datasets, where criteria and scoring guidelines are generated from evaluation judgments. 
This expansion of our dataset enhances generalization capabilities without undermining the existing strengths of the evaluation model.

We conduct a comprehensive examination of Fennec across various evaluation tasks, encompassing both pairwise-eval and single-eval, and scenarios such as Auto-P, PandaLM test set, and MT-bench benchmarks. 
The results emphasize the superiority of our method over recent endeavors in evaluation modeling, showcasing competitive performance comparable to GPT-4 on some specific datasets.
Furthermore, our model excels in refining responses with low scores, leading to enhanced performance in MT-Bench.
In summary:
\begin{enumerate}
    \item[(\uppercase\expandafter{\romannumeral1})] We propose a step-by-step evaluation framework Fennec by incorporating branching and bridging techniques. This initiative aims to replace human evaluation across a spectrum of conversational tasks with complex intents.
    \item[(\uppercase\expandafter{\romannumeral2})] We released a novel open-source training dataset and a freely accessible evaluation model for training and inference with Fennec.
    \item[(\uppercase\expandafter{\romannumeral3})] Our extensive experiments and analyses provide fresh insights for future evaluation endeavors, tackling aspects like the establishment of fine-grained evaluation. 
\end{enumerate}

These contributions pave the way for the development of automated evaluation techniques and the construction of private models.

\begin{figure*}[t]
    \centering
    \includegraphics[scale=0.45]{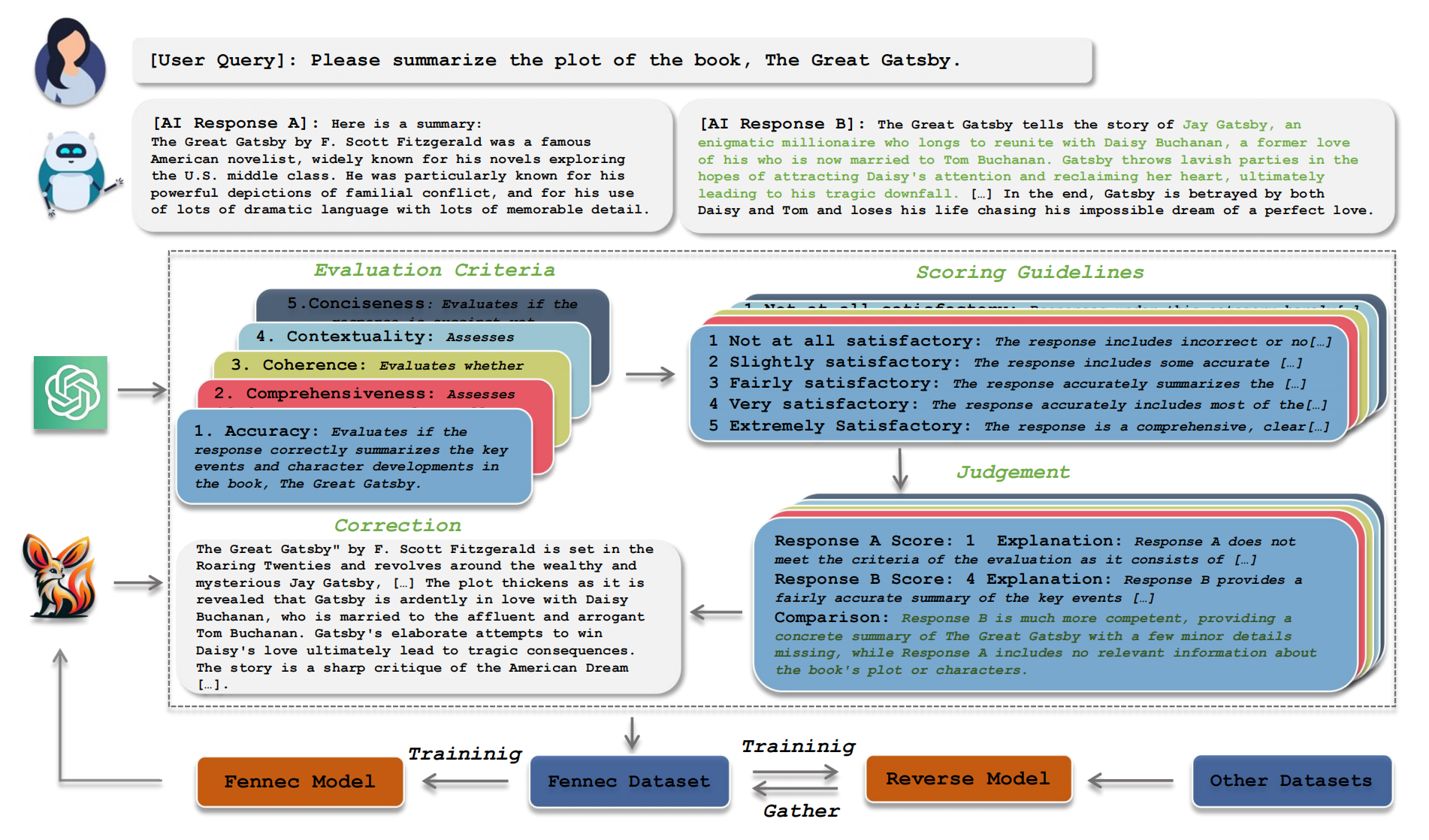}
    \caption{This Figure illustrates Fennec to handle two conversation responses. 
    We systematically define the dimensions of evaluation and scoring granularity to construct a context for the dialogue response. 
    Subsequently, we derive judgments and correction outcomes. More detailed output results can be referenced in the Appendix~\ref{sec:app_detail}.}
    \label{fig:overview}
    \vspace{-0.3cm}
\end{figure*}

\section{Related Work}

With significant advancements in large language models across diverse open-end generation tasks, traditional evaluation metrics such as BLEU~\cite{papineni2002bleu}, ROUGE~\cite{rouge2004package},  BERTScore~\cite{zhang2019bertscore}, and GPTScore~\cite{fu2023gptscore} no longer sufficiently capture the nuanced performance gaps among various models in these tasks.
As a result, Chatbot Arena~\cite{zheng2023judging} has emerged as a widely embraced artificial evaluation platform, utilizing Elo rating to rank different open-source models.
In addition to human evaluation methods, API-based approaches, exemplified by GPT-4~\cite{openai2023gpt4}, have demonstrated the ability to yield judgments closely aligned with those of human experts.
However, these methods are plagued by prohibitive costs, uncontrolled versioning, closed-source nature~\cite{kim2023prometheus}, and potential risks of data leakage and performance instability.
To mitigate these challenges, initiatives such as PandaLM~\cite{wang2023pandalm}, JudgeLM~\cite{zhu2023judgelm}, Auto-J~\cite{li2023generative}, etc., have endeavored to leverage open-source models and collected evaluation data to train dedicated evaluation models.
The consistency between these models and human evaluations on specific datasets can either match or surpass that achieved by GPT-4.
Our research methodology aligns with mainstream practices, primarily concentrating on evaluation tasks that involve more universal and complex conversational data.  
Additionally, certain task-oriented and multi-agent evaluation methods~\cite{jiang2023tigerscore, liu2023x, chan2023chateval}, owing to their specificity, fall outside the current scope of this work. Nonetheless, it is noteworthy that Fennec can naturally adapt to these tasks and scenarios.


In particular, \citet{wang2023pandalm} employed human annotated dataset to fine-tune large language models, leading to the development of an evaluation model.
~\citet{li2023generative} constructed an evaluation dataset covering 58 scenarios through heuristic filtering strategies and post-processing methods and subsequently trained a 13B evaluation model with it.
~\citet{zhu2023judgelm} established a training dataset of 100k instances and focused on designing various training methods, such as swap augmentation, reference support, and reference drop, to alleviate bias problems within the evaluation model during training.
Works most similar to ours include Branch-Solve-Merge (BSM)~\cite{saha2023branch} and Prometheus~\cite{kim2023prometheus}, both of which also decompose the evaluation process. 
However,~\citet{saha2023branch} only propose a task decomposition method without addressing specific inconsistencies, such as evaluation dimensions and scales.
\citet{kim2023prometheus} decompose the evaluation task during data construction but adopt a one-step inference strategy, limiting its evaluation capabilities.
Our approach takes into account the strengths of previous methodologies, incorporating both branch reasoning and multi-step generation.
What is particularly noteworthy is its commitment to maintaining consistency between training and inference, supporting various evaluation scenarios.

\section{Approach}

We leverage the dataset $\mathcal{D}_{\texttt{train}}$ $= \{ \texttt{Prompt}, \mathcal{X}, \mathcal{Y} \}$ for training the evaluation model, where each sample consists of a specific \texttt{Prompt}, a paired input $\mathcal{X}$, and the correlated output $\mathcal{Y}$.
As shown in Figure~\ref{fig:overview}, we offer a thorough explanation of the \textbf{branching} operation with a focus on dataset construction and the \textbf{bridging} operation to consolidate multiple datasets used for training the evaluation model.

\subsection{Branching}
Initially, we utilize GPT-4 model $\mathcal{P}_{\texttt{G4}}$ as a substitute for human annotation to generate the aforementioned dataset $\mathcal{D}_{\texttt{train}}$, which encompasses four sub-datasets 
$ \{ \mathcal{D}_{\texttt{criteria}}, \mathcal{D}_{\texttt{score}}, \mathcal{D}_{\texttt{judge}}, \mathcal{D}_{\texttt{correction}}\}$. 
It is noteworthy that our evaluation model, during the inference, will adopt the same procedural steps as the data construction process, commonly denoted as \textit{behavior cloning}~\cite{ouyang2022training}.

Given the paired user query and their corresponding AI assistant response data $ \{ \mathcal{Q}_{\texttt{user}}, \mathcal{R}_{\texttt{AI}} \}$, we first generate criteria datasets $\mathcal{D}_{\texttt{criteria}} = \{ \mathcal{X}_\texttt{c}, \mathcal{Y}_{\texttt{c}} \}$ across multiple evaluation dimensions for each query.
In this context, the incorporation of multiple dimensions aims to expand candidate space, enabling thorough exploration and facilitating the identification of more rational evaluation~\cite{saha2023branch}.
We specify the maximum dimensions $N$ to control the candidate spaces:
\begin{equation}
\begin{aligned}
    \mathcal{Y}_\texttt{c} \sim \mathcal{P}_{\texttt{G4}}(\mathcal{Y} | \texttt{Prompt}_\texttt{c}, \mathcal{X}_{\texttt{c}}), 
\end{aligned}
\end{equation}
where $\mathcal{X}_{\texttt{c}} \in \{ \mathcal{Q}_{\texttt{user}} \}$  and $\texttt{Prompt}_\texttt{c}$ serve as the language instruction to guide the language model to generate appropriate outputs.

The scoring guidelines can serve the purpose of offering a more fine-grained specification for diverse responses, enabling subtle distinctions between different responses.
It is crucial to finely design scoring guidelines $\mathcal{D}_{\texttt{score}} = \{ \mathcal{X}_\texttt{s}, \mathcal{Y}_{\texttt{c}}, \mathcal{Y}_{\texttt{s}} \}$ to circumscribe the model's decision:
\begin{equation}
\begin{aligned}
    \mathcal{Y}_\texttt{s} \sim \mathcal{P}_{\texttt{G4}}(\mathcal{Y} | \texttt{Prompt}_\texttt{s}, \mathcal{X}_{\texttt{s}}, \mathcal{Y}_\texttt{c}), 
\end{aligned}
\end{equation}
where $\mathcal{X}_{\texttt{s}} \in \{ \mathcal{Q}_{\texttt{user}} \}$  and $\texttt{Prompt}_\texttt{s}$ guide the model $\mathcal{P}_{\texttt{G4}}$ to generate detailed scoring guidelines ranging from 1 to 5 for the current sample.

Based on the established criteria and scoring guidelines, the final evaluation judgments $\mathcal{D}_{\texttt{judge}}$ $= \{ \mathcal{X}_\texttt{j}, \mathcal{Y}_{\texttt{j}}, \mathcal{Y}_{\texttt{c}}, \mathcal{Y}_{\texttt{s}} \}$ can be derived:
\begin{equation}
\begin{aligned}
    \mathcal{Y}_\texttt{j} \sim \mathcal{P}_{\texttt{G4}}(\mathcal{Y} | \texttt{Prompt}_\texttt{j}, \mathcal{X}_{\texttt{j}}, \mathcal{Y}_\texttt{c}, \mathcal{Y}_\texttt{s}),
\end{aligned}
\end{equation}
where $\mathcal{X}_{\texttt{j}} \in \{ (\mathcal{Q}_{\texttt{user}} , \mathcal{R}_{\texttt{AI}}) \}$.
The $\texttt{Prompt}_\texttt{j}$ can be classified into two categories: \textbf{\textit{single-eval}} and \textbf{\textit{pairwise-eval}}.
Single-evaluation primarily addresses scenarios where only a singular response necessitates evaluation.
Meanwhile, when faced with scenarios demanding judgment between two responses, we opt for the pairwise-evaluation.
The distinction between the two manners lies in the fact that pairwise-eval considers two responses, thereby establishing mutual references that enhance the accuracy of scoring.
Please note that our new dataset is only constructed using the pairwise-evaluation method and achieves the ability for single-evaluation through a bridging operation.
It is noteworthy that the criteria and scoring guidelines provide explicit guidance on scoring considerations, thereby simplifying the task of predicting judgments and minimizing errors.

We aggregate all the evaluation judgements $\mathcal{\widehat{Y}}_{j} = \sum_{i=1}^{N} \mathcal{Y}_{\texttt{j}, i}$ across all $N$ dimensions to generate correction suggestion datasets $\mathcal{D}_{\texttt{correction}}$  $ = \{ \mathcal{X}_\texttt{corr}, \mathcal{Y}_{\texttt{corr}}, \mathcal{\widehat{Y}}_\texttt{j} \}$, which will be employed to make focused refinements:
\begin{equation}
\begin{aligned}
    \mathcal{Y}_\texttt{corr} \sim \mathcal{P}_{\texttt{G4}}(\mathcal{Y} | \texttt{Prompt}_\texttt{corr}, \mathcal{X}_{\texttt{corr}}, \mathcal{\widehat{Y}}_\texttt{j}),
\end{aligned}
\end{equation}
where $\mathcal{X}_{\texttt{corr}} \in \{ (\mathcal{Q}_{\texttt{user}}, \mathcal{R}_{\texttt{AI}}) \}$.
Every AI response receiving a low score (scores below 3, as specified in the paper) requires correction to attain a more objective and constructive reply.

The aforementioned evaluation process encompasses four distinct tasks, and we aggregate all these data $\mathcal{D}_{\texttt{train}}$ to fine-tune an open-source pre-trained LLM $\mathcal{P}_{\theta}$ through autoregressive modeling: 
\begin{equation}
\begin{aligned}
    \mathcal{Y} \sim \overrightarrow{\mathcal{P}_{\theta}}(\mathcal{Y} | \texttt{Prompt}, \mathcal{X}).
\end{aligned}
\end{equation}
Following the training, in the evaluation phase, we systematically execute the aforementioned procedures to generate evaluation judgments and refine AI responses. All prompts and specific details are available for reference in the Appendix~\ref{sec:app_prompt}.

\subsection{Bridging}

Increasing the number of tasks and improving data quality~\cite{wei2021finetuned, chung2022scaling} has proven effective in guiding LLM toward better alignment with expected behaviors. 
We elaborate on the integration of additional datasets to expand both quantity and usage scenarios.

Generally, the majority of existing training datasets of evaluation models provide judgment results $\mathcal{Y}_\texttt{j}$ but lack comprehensive evaluation criteria $\mathcal{Y}_\texttt{c}$ and scoring guidance $\mathcal{Y}_\texttt{s}$. 
To address this, we train a reverse model $\overleftarrow{\mathcal{P}_{\theta}}$ using the previously constructed evaluation dataset $\mathcal{D}_{\texttt{train}}$. This model will generate the missing conditions based on judgment results:
\begin{equation}
\begin{aligned}
    \mathcal{\overline{Y}}_\texttt{c} &\sim \overleftarrow{\mathcal{P}_{\theta}}(\mathcal{Y} | \texttt{Prompt}_{rc}, \mathcal{X}_\texttt{r}, \mathcal{Y}_\texttt{j}), \\
    \mathcal{\overline{Y}}_\texttt{s} &\sim \overleftarrow{\mathcal{P}_{\theta}}(\mathcal{Y} | \texttt{Prompt}_{rs}, \mathcal{X}_\texttt{r}, \mathcal{\overline{Y}}_\texttt{c}, \mathcal{Y}_\texttt{j}),
\end{aligned}
\end{equation}
where $\mathcal{X}_{\texttt{r}} \in \{ (\mathcal{Q}_{\texttt{user}}, \mathcal{R}_{\texttt{AI}}) \}$.
We utilize $\texttt{Prompt}_{rs}$ and $\texttt{Prompt}_{rc}$ as instructions to ensure that the model outputs are correlated with both the dialogue response and the judgment.
The new judgment datasets $\mathcal{D}_{\texttt{judge}}$ $= \{ \mathcal{X}_\texttt{j}, \mathcal{Y}_{\texttt{j}}, \mathcal{\overline{Y}}_{\texttt{c}}, \mathcal{\overline{Y}}_{\texttt{s}} \}$ can be incorporated into the dataset $\mathcal{D}_{\texttt{train}}$ to train the evaluation model $\overrightarrow{\mathcal{P}_{\theta}}$. 
To ensure the effectiveness of the training process, we filter out behaviorally inconsistent data, thereby ensuring high quality. 

This approach can be extended to novel scenarios through the standardization of task and prompt formats. 
For example, as mentioned earlier, we do not specifically construct training data for single-eval. 
Instead, we integrate datasets with single-eval capabilities through a bridging mechanism.



\section{Implementation}


In this section, we present a detailed implementation of how to construct training datasets for evaluation models. 
Specifically, we employ the \textit{branching} method to construct \textbf{$\mathcal{D}_{\texttt{Fennec}}$}, subsequently utilizing \textit{bridging} techniques to establish \textbf{$\mathcal{D}_{\texttt{Fennec-bridging}}$}.


\subsection{Training Dataset Construction}

For the currently available open-source datasets, there are no datasets suitable for training models using a branching workflow, as illustrated in Table~\ref{tab:data_count}.
Drawing inspiration from the latest release, specifically the conversational dataset\footnote{\url{https://github.com/GAIR-NLP/auto-j/blob/main/data/training/pairwise_traindata.jsonl}} in the Auto-J~\cite{li2023generative}, we have leveraged the capabilities of GPT-4 to systematically regenerate data for each stage within the workflow, as \textbf{$\mathcal{D}_{\texttt{Fennec}}$}. 
This dataset comprises a total of 57k instances, including 3,314 entries for the formulation of evaluation criteria ($N=5$), where additional comprehensive data statistical details can be found in Appendix~\ref{sec:app_stat}.

In recent open-source evaluation endeavors~\cite{li2023generative, zhu2023judgelm, kim2023prometheus}, a prevailing practice is the direct presentation of judgments, notably lacking associated criteria and scoring guidance.
We employ bridging techniques to systematically construct these missing information.
\textbf{(1)} \textbf{$\mathcal{D}_{\texttt{Auto-J}}$}: Auto-J comprises 3,436 paired training samples, encompassing a range of tasks such as summarization, exam questions, and code analysis, spanning 58 diverse scenarios. 
Each scenario consists of 100 individual samples. 
We utilized a reverse model $\overleftarrow{\mathcal{P}_{\theta}}$ to generate specific conditions for these samples. Considering that judgments may involve various specifications, the resulting scoring guidance could potentially encompass multiple evaluation criteria, thereby highlighting the model's capacity for generalization beyond strict adherence to the training data.
\textbf{(2)} \textbf{$\mathcal{D}_{\texttt{JudgeLM}}$}: JudgeLM\footnote{\url{https://huggingface.co/datasets/BAAI/JudgeLM-100K}} comprises 100k evaluation samples, derived from responses provided by 11 LLMs. 
We filter out inappropriate samples in the dataset that require high-quality reference answers to achieve scores.
\textbf{(3)} \textbf{$\mathcal{D}_{\texttt{Prometheus}}$}: Prometheus\footnote{\url{https://huggingface.co/datasets/kaist-ai/Feedback-Collection}} consists of 1k fine-grained score rubrics, 20K instructions, and 100K responses and language feedbacks generated by GPT-4.
We selectively sample a portion of the data to incorporate into the training dataset, rather than utilizing the entirety of the available data.

In the pairwise evaluation, variations in the positions of two different AI responses may lead to inconsistencies in output results. 
This phenomenon is commonly referred to as \textbf{\textit{positional bias}}~\cite{wang2023large, zheng2023judging}. 
To address this concern, we have proactively employed a common practice: using data augmentation techniques, specifically by exchanging the order of responses during training. 
Similarly, we will employ this method to filter out samples from the training data exhibiting inconsistent predicted results.

\begin{table}[]
\centering
\scalebox{0.8}{
\begin{tabular}{l| cccc}
\toprule
\textbf{Dataset} & $\mathcal{D}_{\texttt{criteria}}$ & $\mathcal{D}_{\texttt{score}}$ & $\mathcal{D}_{\texttt{judge}}$ & $\mathcal{D}_{\texttt{correction}}$ \\
\midrule
\texttt{Auto-J} & \ding{55} & \ding{55} & \ding{51} & \ding{55} \\
\texttt{JudgeLM} & \ding{55} & \ding{55} & \ding{51} & \ding{55} \\
\texttt{Prometheus} & \ding{51}$^*$ & \ding{51}$^*$ & \ding{51} & \ding{55} \\
\midrule
\textbf{$\mathcal{D}_{\texttt{Fennec}}$} & \ding{51} & \ding{51} & \ding{51} & \ding{51} \\
\bottomrule
\end{tabular}
}
\caption{The datasets used for training the evaluation model encompass diverse task-related data scenarios. ( \ding{51}$^*$ signifies that the original work did not encompass the generation of these options during inference.)}
\label{tab:data_count}
\vspace{-0.2cm}
\end{table}

\subsection{Training and Inference}

\begin{table*}[t]
\centering
\scalebox{0.75}{
\begin{tabular}{l | c | c | cc | cc | cc}
\toprule
\multicolumn{1}{c|}{\multirow{2}{*}{\textbf{Methods}}} & \multirow{2}{*}{\textbf{Size}} & \multirow{2}{*}{\textbf{Dataset}} & \multicolumn{2}{c|}{\textbf{\texttt{Auto-P}}} & \multicolumn{2}{c|}{ \textbf{\texttt{PandaLM-test}} }   & \multicolumn{2}{c}{ \textbf{\texttt{MT-Bench}} }  \\
\multicolumn{1}{c|}{} & & & \textbf{Agreement$\uparrow$}  & \textbf{Consistency$\uparrow$} & \textbf{Agreement$\uparrow$} & \textbf{Consistency$\uparrow$} & \textbf{Agreement$\uparrow$} & \textbf{Consistency$\uparrow$} \\
\midrule
\multicolumn{9}{l}{\textbf{\textit{ \small Open-source/Closed-source Chat Models}}} \\
\midrule
GPT-4 & - & - & \textbf{62.28} & 86.28 & 60.06 & 74.67 & 46.71 & 67.26 \\
GPT-3.5  & - & - & 42.74 & 62.43 & - & - & - & - \\
Zephyr  & 7B & - & 31.15 & 57.21 & 48.42 & 68.16 & 31.38 & 54.98 \\
\midrule
\multicolumn{9}{l}{\textbf{\textit{ \small Trained Evaluation Models}}} \\
\midrule
Zephyr  & 7B & \textbf{$\mathcal{D}_{\texttt{Auto-J}}$} & 51.67 & 82.06 & 60.26 & 79.28 & 52.82 & 84.19 \\
Zephyr  & 7B & \textbf{$\mathcal{D}_{\texttt{JudgeLM}}$} & 46.98 & 84.48 & 44.54 & 85.28 & 46.35 & 86.26 \\
PandaLM & 7B & \texttt{PandaLM} & 39.44 & 66.88 & - & - & - & - \\
JudgeLM & 7B & \texttt{JudgeLM} &  46.96  & 72.30 & 58.26 & 76.08 & 44.25 & 68.66 \\
Auto-J & 13B & \texttt{Auto-J} & 54.96 & 83.41 & 60.68 & 78.64 & 50.41 & 77.84 \\
\midrule
\multicolumn{9}{l}{\textbf{\textit{ \small Ours}}}    \\
\midrule
\textbf{Fennec} & 7B & \textbf{$\mathcal{D}_{\texttt{Fennec}}$} & 55.80 & 85.52 & 70.67 & 87.89 & 55.24 & 86.15 \\
& 7B & \textbf{$\mathcal{D}_{\texttt{Fennec-bridging}}$} & 57.40 & \textbf{87.00} & \textbf{72.17} & \textbf{89.69} & 55.65 & \textbf{86.73} \\
\bottomrule
\end{tabular}
}
\caption{The main results on three benchmarks, evaluated through the pairwise-eval, bolded numbers represent the current best results. (Due to the frequent format errors in GPT-3.5 results, some experiments did not include it.)}
\label{tab:pairwise}
\vspace{-0.3cm}
\end{table*}

We utilize Zephyr-7B Chat\footnote{\url{https://github.com/huggingface/alignment-handbook}}~\cite{tunstall2023zephyr}, an aligned version of Mistral-7B~\cite{jiang2023mistral}, as the backbone to train an evaluation model.
We train our model with three epochs on $8$ A100s (40GB).
We use a cosine learning rate scheduler with a peak learning rate of $1e-5$ and $10\%$ warmup steps. 
We train all models with a global batch size of $512$ and use packing with a sequence length of $2,048$ tokens.
It is noteworthy that, despite its diverse origins, the data within \textbf{$\mathcal{D}_{\texttt{Fennec-bridging}}$} has been standardized into a uniform training format, thereby facilitating the utilization of fine-tuning frameworks for training.
During inference, we do not adopt a sampling approach for correction; instead, other processes utilized a temperature setting of $0$ for result generation.

\section{Experiments}

\subsection{Pairwise Evaluation}
We leverage three benchmarks to compare the performance of different evaluation models under the pairwise evaluation setting by comparing two distinct AI responses of different models: Auto-P~\cite{li2023generative}, PandaLM\footnote{\url{https://github.com/WeOpenML/PandaLM}}~\cite{wang2023pandalm}, and MT-bench\footnote{\url{https://github.com/lm-sys/FastChat/blob/main/fastchat/llm_judge/README.md}}~\cite{zheng2023judging} (Due to the model's paradigm and contextual constraints, our evaluation specifically focused on the first round of results).
To emphasize the efficacy of our approach, we have selected recent works as baselines, including PandaLM, JudgeLM~\cite{zhu2023judgelm}, and Auto-J, which have been trained using specific training datasets for evaluation. 
Additionally, we introduce untrained baselines like Zephyr, alongside robust commercial models GPT-3.5 and GPT-4.
Furthermore, we employ \textbf{Consistency} after exchanging response orders and \textbf{Agreement} in consistent assessments aligning with human judgments as the fundamental metrics for the current task.
Particularly noteworthy is the fact that the agreement metric does not represent the accuracy of all data. 
Instead, it indicates the accuracy of model predictions when there is consistency.

As depicted in Table~\ref{tab:pairwise}, the results demonstrate that our model, trained on the \textbf{$\mathcal{D}_{\texttt{Fennec}}$} dataset, consistently outperforms existing open-source evaluation models across multiple benchmarks. 
This highlights the efficacy of the branching method in improving a model's judgment capability on conversational data. 
Furthermore, the Fennec not only surpasses Auto-J on the \texttt{Auto-P} dataset but also maintains performance across other datasets, emphasizing the effectiveness of our approach.
It indicates a substantial improvement in the model's generalization capacity through multiple dimensional and granular evaluations. 
Additionally, the utilization of the \textbf{$\mathcal{D}_{\texttt{Fennec-bridging}}$} dataset further enhances model performance, substantiating the scalability of the bridging method.
Furthermore, we compared the performance under the same backbone model, Zephyr, training with different datasets, such as training with filtered $\mathcal{D}_{\texttt{Auto-J}}$ and $\mathcal{D}_{\texttt{JudgeLM}}$. 
The results indicate that these bridging datasets do not surpass the performance of Fennec. 
Through this comparison, we not only validate the effectiveness of our data construction and inference approach but also demonstrate that bridging can harmonize the advantages of multiple datasets to further enhance the evaluation model.

\subsection{Single Evaluation}

For single evaluation, the task involves assigning a reasonable score to an individual response without relying on any references. 
While the absence of references may lead to a performance reduction, there are still numerous real-world scenarios where evaluating individual responses is necessary.
In these settings, there is a widespread absence of human-annotated evaluation datasets, and the lack of a unified domain and granularity in scoring makes evaluation challenging.
In order to effectively test the consistency between model judgment results and human behavior, we continue to employ three benchmarks similar to pairwise evaluation settings.
Diverging from previous evaluation metrics, single-eval doesn't require swapping the order of responses. 
Hence, we employ \textbf{Accuracy} as the metric to validate the model.

\begin{table}[]
\small
\scalebox{1.0}{
\begin{tabular}{l| ccc}
\toprule
\textbf{Methods} & \textbf{\texttt{Auto-P}} & \textbf{\texttt{PandaLM-test}} & \textbf{\texttt{MT-Bench}} \\
\midrule
\multicolumn{4}{l}{\textbf{\textit{ \small Pairwise-Eval}}}    \\
\midrule
Auto-J & 60.28 & 69.44 & 58.08 \\
PandaLM & - & 59.26 & - \\
\textbf{Fennec} & \textbf{61.84} & \textbf{76.91} & \textbf{60.60} \\
\midrule
\multicolumn{4}{l}{\textbf{\textit{ \small Single-Eval}}}    \\
\midrule
Prometheus & 47.09 & 45.54 & 47.89 \\
\textbf{Fennec} & \textbf{58.15} & \textbf{61.46} & \textbf{58.85} \\
\bottomrule
\end{tabular}
}
\caption{The results evaluated through single-eval, employing \textbf{Accuracy} as the evaluation metric.}
\label{tab:single}
\end{table}

As shown in Table~\ref{tab:single}, it reveals a significant superiority of our approach over the Prometheus, even surpassing certain reference-based evaluation models like PandaLM. 
One major reason is that our method can provide clearer evaluation rules, which aid in implicit alignment across different evaluations, resulting in more accurate scoring.
Simultaneously, the results also demonstrate that the bridging approach can effectively decompose various evaluation processes, facilitating seamless extension to new scenarios.
Besides, we present the Accuracy results of pairwise-eval, highlighting its performance superiority over single-eval.
Such results align with previous research findings: reference items play a pivotal role in evaluations, indicating the preference for employing pairwise evaluation in practical applications.

\subsection{Response Correction}

\begin{table}[]
\centering
\small
\begin{tabular}{l| c | c}
\toprule
\textbf{Models} & \textbf{\texttt{MT-Bench}} & \textbf{Impr $\uparrow$}\\
\midrule
GPT-4 & \textbf{8.96} & - \\
WizardLM-30B & 7.13 & - \\
LLaMA2-13B Chat & 7.06 & - \\
LLaMA2-70B Chat & 6.99 & - \\
\midrule
LLaMA2-7B Chat & 6.26 & - \\
\textbf{w/ Fennec Correction} & \textbf{7.15} & + 0.89 \\
\midrule
Alpaca-13B & 4.97 & - \\
\textbf{w/ Fennec Correction} & \textbf{6.84} & + 1.87 \\
\bottomrule
\end{tabular}
\caption{The main results on the MT-bench, refined through Fennec correction to enhance the chat model's output. (Impr signifies the performance improvement.)}
\label{tab:correction}
\vspace{-0.3cm}
\end{table}

To test the model's performance in response correction, we conducted an evaluation and correction (if necessary) using the MT-bench dataset.
In contrast to the aforementioned dataset, this dataset comprises 160 questions spanning eight distinct domains of knowledge (we solely utilized results from the first turn).
Specifically, we employed the output of Alpaca~\cite{alpaca} and LLaMA2 Chat~\cite{touvron2023LLaMA} models as the dialogues requiring evaluation, and subsequently utilized Fennec for evaluation and correction.
We employ GPT-4 to assign scores ranging from 1 to 10~\cite{zheng2023judging} and also include the results of WizardLM~\cite{xu2023wizardlm} for comparison.

The results are shown in Table~\ref{tab:correction}, and it is evident that the responses after correction exhibit a significant performance improvement. LLaMA2-7B Chat achieved an improvement of 0.89 points, and Alpaca showed a notable increase of 1.87 points. 
Notably, for the lower-performing LLM assistant, a more substantial enhancement is observed compared to stronger models. 
This emphasizes the critical role of our Fennec correction step in achieving satisfactory responses, particularly for responses with lower performance. 
Moreover, the results demonstrate that after refinement, the model performance of LLaMA2-7B Chat even surpasses that of LLaMA2 13B and 70B Chat models, highlighting the effectiveness of our approach.

\section{Discussions and Analysis}

\subsection{Exploring the Effectiveness of Branching}

\begin{figure}
    \centering
    \includegraphics[scale=0.5]{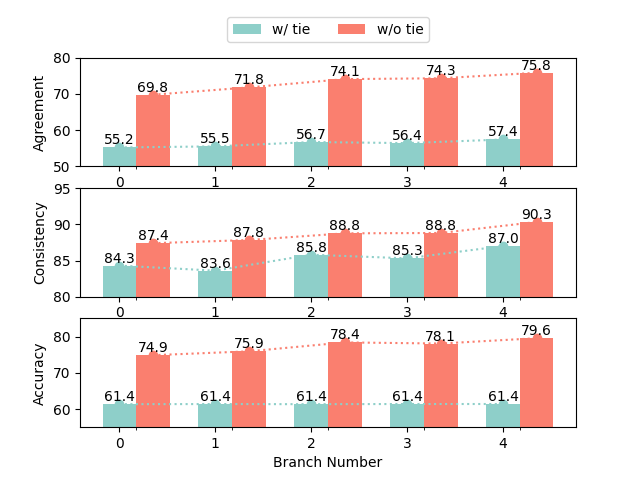}
    \caption{The Figure illustrates the performance of Fennec on Auto-P with varying branch numbers.}
    \label{fig:branch}
    \vspace{-0.3cm}
\end{figure}

In this section, we delve into the examination of the influence of varying branch numbers on performance evaluation. 
As presented in Table~\ref{fig:branch}, the results encompass diverse branch numbers, ranging from 1 to 5, on Auto-P (other results refer to Appendix~\ref{sec:app_branch}). 
The results reveal an ascending trend in both Agreement and Consistency. However, in contrast, Accuracy demonstrates minimal to no variation.
For a deeper understanding, we carried out a recalibration experiment on the model's performance metrics, excluding samples labeled as ``tie''. 
This analysis revealed an improvement in consistency across all metrics.

From the observations, we can conclude that \textbf{\textit{the branching strategy proven more effective in distinguishing disparities among various dialogue responses, enabling the identification of a clear winner rather than settling for a tied judgment.}}
Nevertheless, the increase in branch numbers may not authentically reflect changes in model performance, particularly in datasets abundant with tied samples. 
Therefore, an evaluation exclusively on samples labeled as ``win/los'' becomes essential.
It is important to note that instances labeled as ``tie'' represent a compromise made by human annotators encountering challenges in identifying more suitable evaluation dimensions.
Our approach adeptly addresses this limitation and is particularly suitable for accurately distinguishing responses in various scenarios.
This flexibility allows for tailored adjustments to mitigate its shortcomings.

\subsection{The Impact of Backbone Model}

\begin{table}[]
\centering
\small
\scalebox{0.88}{
\begin{tabular}{l| ccc}
\toprule
\multicolumn{1}{c|}{\multirow{2}{*}{\textbf{Backbone}}} & \multicolumn{2}{c}{\textbf{\texttt{Auto-P}}} \\
\multicolumn{1}{c|}{} &
\textbf{Agreement$\uparrow$}  & \textbf{Consistency$\uparrow$} & 
\\
\midrule
LLaMA2-7B Chat & 44.68 & 86.67 \\
Mistral-7B & 54.49 & 84.25 \\
\midrule
\textbf{Zephyr-7B} & 55.80 & 85.52 \\
\bottomrule
\end{tabular}
}
\caption{The results for different backbone models.}
\label{tab:backbone}
\vspace{-0.5cm}
\end{table}

To investigate how different backbone models affect evaluation datasets, we utilized the \textbf{$\mathcal{D}_{\texttt{Fennec}}$} dataset to train various baseline models.
In contrast to the Zephyr used in previous studies, we also tested models such as Mistral-7B~\cite{jiang2023mistral} (without alignment to the preference dataset) and the LLaMA2-7B Chat model. 
The results presented in Table~\ref{tab:backbone} illustrate the performance of these evaluation models on \texttt{Auto-P}. 
The findings indicate that Fennec, trained with Zephyr, outperforms all other models.
By contrasting these results with Mistral, the aligned model demonstrates superior performance compared to the original model.
This emphasizes the notion that aligning the backbone models with human-like behavioral capabilities leads to significant improvements in the evaluation models' performance.
This phenomenon also encourages us to train the evaluation model in the future to preferentially utilize aligned models.

\subsection{System-level Ranking}

\begin{figure}
    \centering
    \vspace{-0.5cm}
    \includegraphics[scale=0.45]{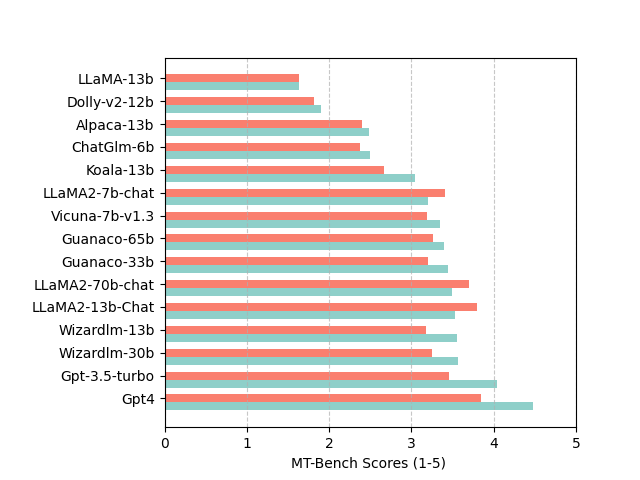}
    \caption{Values and ranking on MT-Bench. (We rescale the GPT-4 scores to 1-5 by multiplying with factor 0.5.)}
    \label{fig:rank}
    \vspace{-0.5cm}
\end{figure}

As an evaluation model, a crucial attribute is the ability to rank different models, especially for comparing the performance of models trained with different settings or datasets. 
We gathered output results from diverse open-source LLMs, including LLaMA2, Dolly~\cite{DatabricksBlog2023DollyV1}, ChatGLM~\cite{zeng2022glm}, Koala~\cite{koala_blogpost_2023}, Guanaco~\cite{dettmers2023qlora}, etc.
To achieve the experiment, we utilized single-eval to obtain the final scores and then compared the ranking results with GPT-4.
From the results in Figure~\ref{fig:rank}, it is evident that Fennec aligns well with GPT-4. 
This showcases our model's capacity to discern subtle response differences through multidimensional evaluations, resulting in more accurate judgments. 
Certainly, it is inevitable that evaluating some complex data may be challenging. 
We believe that this phenomenon can be improved by leveraging a larger backbone and more extensive datasets.

\section{Conclusion}

In this work, we present Fennec, a step-by-step framework designed for evaluating conversational response using a \textit{branching} mechanism, facilitating a systematic generation of evaluation results. Additionally, we introduce a novel training dataset and an expansion method called \textit{bridging} to enhance the model's performance and scalability.
Fennec demonstrates remarkable performance and correction capabilities compared to the current state-of-the-art open-source evaluation models across various benchmarks.
In the near future, we envision more works exploring the application of branching mechanisms in fine-grained evaluation tasks. 
Considering its efficacy in discerning subtle differences, we believe this approach holds promise for evaluation in complex conversational scenarios.

\section*{Limitations}
The need to address two current points for improvement in our model is evident. 
Firstly, we employ a branching method for evaluation, unavoidably leading to increased inference time. 
In the current approach, there is an absence of an effective ranking system for the candidate space, necessitating traversal to identify suitable evaluation criteria. 
Therefore, an expected improvement involves efficiently ranking the candidate space through preference training, aiding in the identification of high-priority evaluation criteria within a reasonable time limit. 
Additionally, another aspect to consider is that the current training data and scenarios are limited, allowing evaluation only for content related to single-turn dialogues and common scenarios. 
Consequently, in future work, our goal is to expand the range of usage scenarios, encompassing multi-turn dialogues and knowledge-intensive situations such as fact verification and common-sense reasoning.

\section*{Ethics Statement}
One notable aspect is whether our dataset includes content requiring ethical scrutiny.
Issues related to hallucinatory outputs and erroneous content are often encountered in LLMs. However, in the construction of our dataset, we rigorously controlled the sources of dialogue data required and used GPT-4 for content generation. Currently, ChatGPT has strict content moderation, filtering out a significant portion of inappropriate content. 
However, the use of our model for content moderation is still debatable. 
Our model is designed to identify and correct inappropriate model outputs, but currently, such content has not been present in the training data. Of course, we will address this in future work to make better judgments about additional content.


\bibliography{anthology,custom}
\bibliographystyle{acl_natbib}

\clearpage
\appendix
\section{Appendix}

\subsection{The Prompts of Fennec}
\label{sec:app_prompt}
In Table~\ref{tab:app_system}, we initially present the system messages for training Fennec models. 
These messages can set specific prompts for each execution step, aiding the model in generating improved results.
In Table~\ref{tab:app_prompt}, we illustrate the prompts needed for each task in the branching process. 
These instructions assist the model in generating outputs in a particular format. 
Utilizing these specific output formats helps us extract reply scores more effectively through rule-based approaches.
In Table~\ref{tab:app_bridging_prompt}, we have rephrased the evaluation instructions from previous Auto-J and JudgeLM approaches, incorporating evaluation dimensions and scoring rules. 
For Prometheus data, we employed single-evaluation prompts, as previously mentioned.

\subsection{The Details of Overview}
\label{sec:app_detail}
In this section, we present the comprehensive set of data generated throughout the dialogue training process. This encompasses:
1) Crafting evaluation criteria, as delineated in Table~\ref{tab:app_1}.
2) Articulating scoring guidelines, elucidated in Table~\ref{tab:app_1}.
3) Generating judgments, explicated in Tables~\ref{tab:app_2}, \ref{tab:app_3}, \ref{tab:app_4}, and \ref{tab:app_5}.
4) Creating correction content, as detailed in Table~\ref{tab:app_6}.
This meticulous documentation encapsulates the multifaceted aspects involved in the training process and provides a structured overview of the evaluative components employed.

\begin{table}[ht]
\centering
\scalebox{0.88}{
\begin{tabular}{l| c}
\toprule
\textbf{Datasets} & \textbf{Data Number}\\
\midrule
Evaluation Criteria & 3,314 \\
Scoring Guidelines & 16,493 \\
Judgement & 32,986 \\
Correction & 4,846 \\
\midrule
$\mathcal{D}_{\texttt{Fennec}}$ & 57,639 \\
\midrule
$\mathcal{D}_{\texttt{Auto-J}}$ & 6,872 \\
$\mathcal{D}_{\texttt{JudgeLM}}$ & 53,548 \\
$\mathcal{D}_{\texttt{Prometheus}}$ & 16,730 \\
\midrule
$\mathcal{D}_{\texttt{Fennec-bridging}}$  & 134,789 \\
\bottomrule
\end{tabular}
}
\caption{The statistics of training datasets.}
\label{tab:stat}
\vspace{-0.5cm}
\end{table}

\subsection{The Statistics of Training Datasets}
\label{sec:app_stat}

The Table~\ref{tab:stat} presents data statistics of the Fennec training dataset. 
Considering that GPT-4 API may deem some queries in the conversational data unanswerable, these have been excluded from the final dataset.
Notably, within the $\mathcal{D}_{\texttt{Fennec-bridging}}$ dataset, data from $\mathcal{D}_{\texttt{Auto-J}}$, $\mathcal{D}_{\texttt{JudgeLM}}$, and $\mathcal{D}_{\texttt{Prometheus}}$ were subject to filtering to expedite the training process.
In this context, our primary criterion for filtering out redundant data was the consistency in predicted scores.

\begin{figure}
    \centering
    \includegraphics[scale=0.5]{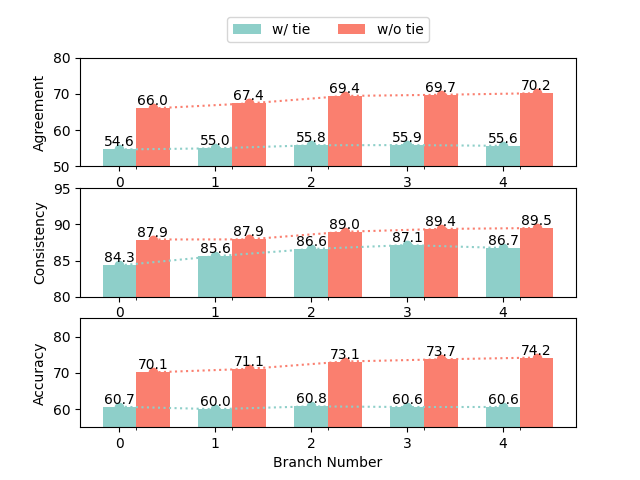}
    \caption{The Figure illustrates the performance of Fennec on MT-bench with varying branch numbers.}
    \label{fig:app_branch1}
    \vspace{-0.3cm}
\end{figure}

\begin{figure}
    \centering
    \includegraphics[scale=0.5]{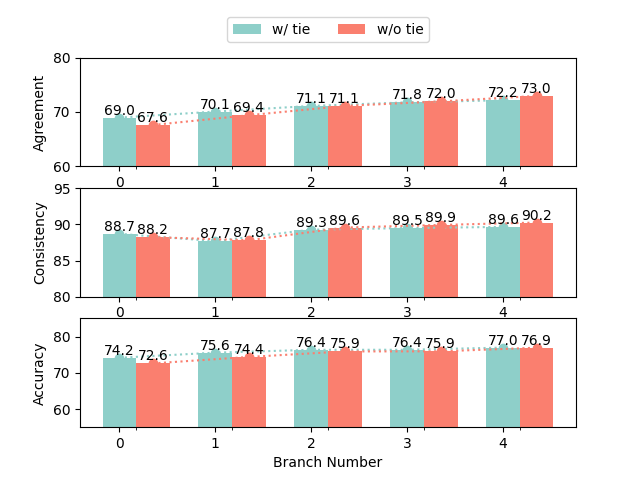}
    \caption{The performance of Fennec on the PandaLM test set with varying branch numbers.}
    \label{fig:app_branch2}
    \vspace{-0.3cm}
\end{figure}

\subsection{Exploring the Effectiveness of Branching}
\label{sec:app_branch}
We present Fennec's performance, manipulating the branch number on both the PandaLM test set and MT-bench, as depicted in Figure~\ref{fig:app_branch1} and Figure~\ref{fig:app_branch2}. 
Remarkably, MT-Bench demonstrates a significant 30\% occurrence of tie labels, surpassing the PandaLM dataset's more modest 10\% frequency.
In situations where ``tie'' label samples are less frequent in the evaluation dataset, our model significantly outperforms GPT-4, especially in evaluations that incorporate multiple branches.
This supports the claim that our model excels in situations requiring discernment in response refinement.

\begin{table*}[t]
\centering
\scalebox{0.85}{
\begin{tabular}{ l | p{14cm} }
\toprule
\textbf{Step} &  \textbf{Content} \\
\midrule
Criterion & You are a fair, faithful, and helpful content evaluation assistant. Kindly assist me in accomplishing the assigned task by creating Evaluation Criteria for the provided dialogue. \\
\midrule
Scoring Guidelines & You are a fair, faithful, and helpful content evaluation assistant. Kindly assist me finalize the assigned task by developing Evaluation Criteria into comprehensive Scoring Guidelines. \\
\midrule
Pairwise-eval &  You are a fair, faithful, and helpful content evaluation assistant. Kindly assist me in finishing the assigned task by providing Pairwise Evaluations for the given dialogue. (Tips: This entails evaluating responses through the comparison of two distinct replies.) \\
\midrule
Single-eval &  You are a fair, faithful, and helpful content evaluation assistant. Please assist me in completing the assigned task by providing Single-Score Evaluations for the given dialogue. (Tips: This involves assessing individual responses independently.) \\
\midrule
Correction & You are an assistant capable of assisting in content modification. It is necessary to correct and refine the dialogue based on User Queries, Responses, and corresponding Evaluation results. \\
\bottomrule
\end{tabular}
}
\caption{System messages for Fennec.}
\label{tab:app_system}
\end{table*}

\begin{table*}[t]
\centering
\scalebox{0.85}{
\begin{tabular}{ l | p{14cm} }
\toprule
\textbf{Step} &  \textbf{Content} \\
\midrule
Auto-J & Given a [User Query], please score the responses from two AI assistants according to the [Evaluation Criteria] and [Scoring Guideline]. Ensure a comparative and objective [Judge Result] based on the evaluation criteria and scoring guideline, aiming to identify deficiencies in the response content. Here are the instructions to assess and compare the two responses:1. Pinpoint the key factors to distinguish these two responses.2. Conclude your comparison by providing a final decision on which response is better, or they are tied. Begin your final decision statement with ``So, the final decision is Response 1 / Response 2 / Tie''. Ensure that your decision aligns coherently with the comprehensive evaluation and comparison you've provided.***[User Query]:{query}***Evaluation Criteria]:{criteria}***[Scoring Guideline]:{scoring}***[The Start of Response 1]:{response1}[The End of Response 1]***[The Start of Response 2]:{response2}[The End of Response 2]***Please return [Judge Result]: \\
\midrule
JudgeLM & Given a [User Query], please score the responses from two AI assistants according to the [Evaluation Criteria] and [Scoring Guideline]. Ensure a comparative and objective assessment based on the evaluation criteria and scoring guideline, aiming to identify deficiencies in the response content. Provide a final score of 1-10 along with relevant explanations.***[User Query]:{query}***[Evaluation Criteria]:{criteria}***[Scoring Guideline]:{scoring}***[The Start of Assistant 1's Response]:{response1}[The End of Assistant 1's Response]***[The Start of Assistant 2's Response]:{response2}[The End of Assistant 2's Response]***Please return [Judge Result]: \\
\bottomrule
\end{tabular}
}
\caption{Bridging prompts for Fennec.}
\label{tab:app_bridging_prompt}
\end{table*}

\begin{table*}[ht]
\centering
\scalebox{0.85}{
\begin{tabular}{ l | p{14cm} }
\toprule
\textbf{Step} &  \textbf{Content} \\
\midrule
Criteria & For evaluating human satisfaction with responses from an AI assistant based on a [User Query], we need to brainstorm and establish five [Evaluation Criteria] directly linked to the user's query. These criteria play a crucial role in objectively assessing response content, with higher priority and greater evaluation weight.***As an illustration:1. Relevance: Evaluate whether the response is directly related to the user's query.2. Criterion: Assess the correctness of the information provided in the response. etc.***[User Query]:{query}***Please return five [Evaluation Criteria]: \\
\midrule
Scoring Guidelines &  Consider a [User Query] and [Evaluation Criteria] for evaluating response satisfaction. Reflect on these criteria and offer a comprehensive [Scoring Guideline] on a scale of 1-5 (1 represents 'Not at all satisfactory' and 5 represents 'Extremely satisfactory'). Ensure that these guidelines are closely tied to both the user query and the assessment criteria, allowing for a precise evaluation of possible responses to the user query. Conduct a detailed comparison of the [Scoring Guideline] to ease adherence and assist individuals in assigning reasonable scores.***[User Query]:{query}***[Evaluation Criteria]:{criteria}***Please return detailed [Scoring Guideline]: \\
\midrule
Pairwise-eval & Given a [User Query], please score the responses (A and B) from two AI assistants according to the [Evaluation Criteria] and [Scoring Guideline]. Ensure a comparative and objective assessment based on the evaluation criteria and scoring guideline, aiming to identify deficiencies in the response content. Provide a final score of 1-5 along with relevant explanations.***[User Query]:{query}***[Evaluation Criteria]:{criteria}***[Scoring Guideline]:{scoring}***[The Start of Response A]:{response1}[The End of Response A]***[The Start of Response B]:{response2}[The End of Response B]***Please return [Judge Result] as follows:Response A Score: 3Explanation: Explanation of the score for the Response A.Response B Score: 3Explanation: Explanation of the score for the Response B.Comparison: The comparison of the Response A and Response B.[Judge Result]: \\
\midrule
Single-eval & Given a [User Query], please score the responses from AI assistants according to the [Evaluation Criteria] and [Scoring Guideline]. Ensure a comparative and objective assessment based on the evaluation criteria and scoring guideline, aiming to identify deficiencies in the response content.***[User Query]:{query}***[Evaluation Criteria]:{criteria}***[Scoring Guideline]:{scoring}***[The Start of Response]:{response}[The End of Response]***Assign a score as an integer between 1 and 5. Provide a detailed [Judge Result] strictly based on the given Scoring Guideline, refraining from a general evaluation. Please return [Judge Result] as follows:Response Score: 3Explanation: Explanation of the score for the Response.Please return [Judge Result]: \\
\midrule
Correction & Provided with a [User Query], the AI assistant's [Original Response], and a comprehensive objective evaluation of the response, please attend to the identified shortcomings in the original response according to the [Judge Result]. Make certain that the modified response remains objective, non-harmful, and constructive in addressing the user's query intent, while also aligning with human behavioral norms.***[User Query]:{query}***[The Start of Original Response]:{response}[The End of Original Response]***[The Start of Judge Result]:{judge}[The End of Judge Result]. Kindly return one final [Modified Response] for user query directly without additional information.Please return [Modified Response]:\\
\bottomrule
\end{tabular}
}
\caption{Evaluation prompts for Fennec.}
\label{tab:app_prompt}
\end{table*}

\begin{table*}[]
\centering
\scalebox{0.85}{
\begin{tabular}{ l | p{14cm} }
\toprule
\textbf{Data} &  \textbf{Content} \\
\midrule
User Query  & Please summarize the plot of the book, The Great Gatsby. \\
\midrule
AI Response A  & Here is a summary: The Great Gatsby by F. Scott Fitzgerald was a famous American novelist, widely known for his novels exploring the U.S. middle class. He was particularly known for his powerful depictions of familial conflict, and for his use of lots of dramatic language with lots of memorable detail. \\
\midrule
AI Response B & The Great Gatsby tells the story of Jay Gatsby, an enigmatic millionaire who longs to reunite with Daisy Buchanan, a former love of his who is now married to Tom Buchanan. Gatsby throws lavish parties in the hopes of attracting Daisy's attention and reclaiming her heart, ultimately leading to his tragic downfall. Throughout the story, Gatsby's story of love and loss is contrasted with the hollowness and false values of the wealthy and powerful people that surround him. In the end, Gatsby is betrayed by both Daisy and Tom and loses his life chasing his impossible dream of a perfect love.\\
\bottomrule
\end{tabular}
}
\caption{The exhibited conversational data comprises user query and responses generated by two selected AI.}
\label{tab:app_c}
\end{table*}

\begin{table*}[]
\centering
\scalebox{0.85}{
\begin{tabular}{ p{2cm} | p{14cm} }
\toprule
\textbf{Data} & \textbf{Content} \\
\midrule
Criteria & \textbf{Accuracy}: \textit{Evaluates if the response correctly summarizes the key events and character developments in the book, The Great Gatsby.} \\
\midrule
Scoring Guidelines & \textbf{1. Not at all satisfactory}: \textit{The response includes incorrect or no information about the plot of The Great Gatsby. Major events and character developments are misunderstood or unmentioned. }

\textbf{2. Slightly satisfactory}: \textit{The response includes some accurate information about the plot and characters of The Great Gatsby, but significant elements are omitted or misrepresented. The explanation may lack cohesion or be difficult to follow.}

\textbf{3. Fairly satisfactory}: \textit{The response accurately summarizes the main plot and provides some details about character developments in The Great Gatsby but lacks emphasis on key events. Not all major points are touched upon, or the explanation lacks thoroughness or clarity.}

\textbf{4. Very satisfactory}: \textit{The response accurately includes most of the main plot points and character developments in The Great Gatsby, demonstrating a good understanding of the novel. Some minor elements may be missing, or the explanation may have small inaccuracies.}

\textbf{5. Extremely Satisfactory}: \textit{The response is a comprehensive, clear, and precise summary of the plot of The Great Gatsby, capturing all major events and significant character developments. It reflects a deep understanding and thorough reading of the novel.}

\textbf{In Summary}:
\textit{1 Major inaccuracies and omissions are present. 
2 Some correct information, but many key points are missed or inaccurate.
3 Good summary, although some key events or character developments may be missing.
4 Nearly complete summary with minor omissions or slight inaccuracies.
5 Full, comprehensive summary with perfect accuracy.} \\
\midrule
Judgement & \textbf{Response A Score: 1}

\textbf{Explanation}: \textit{Response A does not meet the criteria of the evaluation as it consists of incorrect information and does not include any details about the book's plot or characters. The explanation given is more about F. Scott Fitzgerald, who is the author of the book and not the actual content of the book. The details provided are incorrect and irrelevant.} 

\textbf{Response B Score: 4}

\textbf{Explanation}: \textit{Response B provides a fairly accurate summary of the key events and characters in The Great Gatsby. It discussed the main character, Jay Gatsby, his love interest Daisy, her husband Tom Buchanan, and the consequences Gatsby faces due to his unrequited love for Daisy. The AI could have included more context and detail about the theme of the American Dream and the time period to secure a score of 5.} 

\textbf{Comparison}: \textit{Response B is much more competent, providing a concrete summary of The Great Gatsby with a few minor details missing, while Response A includes no relevant information about the book's plot or characters.} \\
\bottomrule
\end{tabular}
}
\caption{The scoring guidelines and judgment for the ``Accuracy'' evaluation criterion.}
\label{tab:app_1}
\end{table*}

\begin{table*}[]
\centering
\scalebox{0.85}{
\begin{tabular}{ p{2cm} | p{13cm} }
\toprule
\textbf{Data} & \textbf{Content} \\
\midrule
Criteria & \textbf{Comprehensiveness}: \textit{Assesses if the response encapsulates all important elements in the book, including themes, plot lines, and crucial characters.} \\
\midrule
Scoring Guidelines & \textbf{1. Not at all satisfactory}: \textit{Not at all satisfactory: The response has either not answered the query or only mentioned the name of the book without any further details about its plot, themes, or important characters.}

\textbf{2. Slightly satisfactory}: \textit{The response addresses only a part of the user query, for instance, highlights one or two main events in the plot or mentions one main character, leaving out other significant elements of the story.}

\textbf{3. Fairly satisfactory}: \textit{The response provides a synopsis of the plot and identifies major characters but lacks depth such as nuances in the storyline, interconnectedness of characters, or omission of crucial themes that the book also addresses.}

\textbf{4. Very satisfactory}: \textit{The response provides a comprehensive summary of the plot, identifies all the essential characters, including their contribution to the plot, but only briefly touches upon the thematic elements in the book or their interpretations.}

\textbf{5. Extremely Satisfactory}: \textit{The response provides a thorough summary of the plot, identifies all key characters and explains their role in the story, and discusses major themes in the book like the American Dream, wealth, and social stratification. It also interprets how these themes play out against the backdrop of the 1920's American society, a core element in ``The Great Gatsby''. The response provides an in-depth understanding of the text and stimulates further thinking.}

\textbf{In Summary}:
\textit{In comparing these guidelines, it is crucial to remember that a score of one means that the response is inadequate and fails to satisfy the user's query. In contrast, a score of five means that the response is robust and offers a comprehensive understanding of the book, meeting all the user's expectations. Intermediate scores between 2 to 4 progressively signify the extent of completeness and detail in the response.} \\
\midrule
Judgement &\textbf{Response A Score: 1}

\textbf{Explanation}: \textit{Response A failed to provide a satisfactory summary of ``The Great Gatsby.'' There's only a general background about the author but no specific information related to the plot, characters, or themes of the book, which is the primary requirement of the user's query. This response doesn't meet the necessary criteria for comprehensiveness, as it does not encapsulate any important elements of the book.}

\textbf{Response B Score: 3}

\textbf{Explanation}: \textit{Response B mildly satisfies the query. It has summarized the plot and mentioned the lead characters, i.e., Jay Gatsby, Daisy Buchanan, and Tom Buchanan. However, the response doesn't delve deeper into the nuances of the plotline, interconnectedness of characters, and fails to mention the significant themes of the story. The summary is good but lacks depth and complexity.}

\textbf{Comparison}: \textit{Response B is superior to Response A as it provides specific details about the plot and characters of ``The Great Gatsby,'' partially satisfying the user's query. On the other hand, Response A did not provide any important details about the book aside from a generic reference to the author's other works. Therefore, in terms of comprehensiveness, Response B clearly outmatches Response A.} \\
\bottomrule
\end{tabular}
}
\caption{The scoring guidelines and judgment for the ``Comprehensiveness'' evaluation criterion.}
\label{tab:app_2}
\end{table*}

\begin{table*}[t]
\centering
\scalebox{0.85}{
\begin{tabular}{ p{2cm} | p{13cm} }
\toprule
\textbf{Data} & \textbf{Content} \\
\midrule
Criteria & \textbf{Coherence}: \textit{Evaluates whether the response is logically structured, ensuring that the storyline summarization flows sensibly from the beginning to the end.} \\
\midrule
Scoring Guidelines & \textbf{1. Not at all satisfactory}: \textit{The summary does not reflect the plot of The Great Gatsby. The response is scattered and lacks logical progression. There's no connection stated between the beginning, middle, and end of the story and it's unclear what happens in the book based on the response.}

\textbf{2. Slightly satisfactory}: \textit{The summary minimally reflects the plot of The Great Gatsby, but significant details are missing or misrepresented. The response shows an attempt at logical progression, but is not successful, and might jump around chronologically within the storyline.}

\textbf{3. Moderately satisfactory}: \textit{The summary represents the main plot of The Great Gatsby in a partially clear and coherent way. Major plot points are captured and the storyline shows some discernable organization, but minor details may be overlooked or misrepresented. There may exist gaps in chronological progression.}

\textbf{4. Very satisfactory}: \textit{The summary clearly represents the main plot of The Great Gatsby and most key events are captured accurately. The storyline logically progresses from the beginning to end, although minor details may still be overlooked or slightly misrepresented.}

\textbf{5. Extremely Satisfactory}: \textit{The summary accurately and concisely encompasses the entire plot of The Great Gatsby. It is detailed and thorough, capturing all key events from the beginning to the end in a logically coherent fashion. No major or minor details are overlooked, and the response reflects a clear and complete understanding of the book.}

\textbf{In Summary}:
\textit{the scoring increases from 1 to 5, the summary provided becomes more thorough, accurate, and logically structured. At lower scores, there are large gaps within the summary, misrepresentations of important details, and poor flow, whereas at higher scores the response accurately represents every aspect of the plot in a very detailed, thorough, and logical manner. A score of 1 would indicate a lack of understanding of the story, while a score of 5 would reflect a complete understanding and perfect representation of the plot.} \\

\midrule
Judgement &\textbf{Response A Score: 1}

\textbf{Explanation}: \textit{Response A does not summarize the plot of The Great Gatsby at all. It only provides a general comment on the author's body of work, rather than focusing on the plot of the requested book. It completely fails to mention any storyline or character in the novel, thus not providing any form of summarization, resulting in the lowest score as per the Scoring Guideline.}

\textbf{Response B Score: 4}

\textbf{Explanation}: \textit{Response B provides an accurate overview of the main plot of The Great Gatsby. It covers the central character, Jay Gatsby, his longing for Daisy Buchanan, the parties he throws, and his tragic downfall. However, it fails to mention some minor but vital details such as the role of the narrator, Nick Carraway, and the dichotomy between East Egg and West Egg. Therefore, though very satisfactory, the score cannot reach the highest rank due to slightly missed details.}

\textbf{Comparison}: \textit{Comparatively, Response B provides a clear, logically structured plot summary of The Great Gatsby, but Response A does not address the user query at all. Response B shows understanding and knowledge of the book, lacking a bit in detail. On the other hand, Response A does not reflect any understanding of the book, resulting in a lower score.}\\
\bottomrule
\end{tabular}
}
\caption{The scoring guidelines and judgment for the ``Coherence'' evaluation criterion.}
\label{tab:app_3}
\end{table*}

\begin{table*}[t]
\centering
\scalebox{0.85}{
\begin{tabular}{ p{2cm} | p{13cm} }
\toprule
\textbf{Data} & \textbf{Content} \\
\midrule
Criteria & \textbf{Contextuality}: \textit{Assesses whether the response reflects the contextual background of the book, such as the social and cultural setting in which the story is set.} \\
\midrule
Scoring Guidelines & \textbf{1. Not at all satisfactory}: \textit{Responses under this category barely address the user query. They do not provide an accurate or any summary of ``The Great Gatsby''. Any mention of the social, cultural, or geographic context in which the story is set is incorrect or completely missing.}

\textbf{2. Not very satisfactory}: \textit{These responses do provide a summary of the book, however, they are laden with inaccuracies or missing key points from the novel's plot. They have a slight mention of the social and cultural setting of the book, but the references added are incorrect, vague, or tangential.}

\textbf{3. Moderately satisfactory}: \textit{Responses falling under this category attempt to provide an accurate summary of ``The Great Gatsby''. They may have minor inaccuracies, but overall, the key elements of the plot are included. They also touch upon the contextual background of the book but may fail to provide a detailed view of the social and cultural atmosphere pertinent to the storyline.}

\textbf{4. Very satisfactory}: \textit{These responses deliver an accurate and comprehensive summary of ``The Great Gatsby'', identifying crucial plot points. The responses also include a clear discussion of the social and cultural setting, indicating an understanding of the impact of these factors on the narrative; however, more depth or precision may be required.}

\textbf{5. Extremely Satisfactory}: \textit{Responses in this category not only provide a very accurate, detailed, and comprehensive summary of ``The Great Gatsby'' but also successfully weave in a rich understanding of the social and cultural setting. These responses demonstrate how the contextual atmosphere of the Roaring Twenties contributes to the characters' motivations, decisions, and the plot's progression.}

\textbf{In Summary}:
\textit{One should consider these guidelines fluid rather than strictly dichotomous. Responses might also display characteristics from adjacent categories and should be rated according to the evaluator's best judgment.} \\

\midrule
Judgement &\textbf{Response A Score: 1}

\textbf{Explanation}: \textit{Response A does not provide a summary of ``The Great Gatsby''. It only mentions the author and vaguely discusses his style, missing the user query altogether which asked for a summary of the plot. It has no mention of the social, cultural, or geographic context in which the story is set. The rating thus falls under the Score 1 category of the scoring guideline.}

\textbf{Response B Score: 4}

\textbf{Explanation}: \textit{Response B provides a coherent summary of ``The Great Gatsby'', identifying key plot points such as Gatsby's longing for Daisy, his extravagant parties, and his tragic end. Furthermore, it also hints at the social and cultural setting of the novel by mentioning the 'wealthy and powerful' people and their 'false values.' However, it could have explored the contextual setting, such as the Roaring Twenties, in more depth to score higher. Therefore, it falls under the Score 4 category of the scoring guideline.}

\textbf{Comparison}: \textit{Response B superiorly addresses the user query compared to Response A, providing not only an accurate summary of the book but also reflecting, to a certain degree, the social and cultural setting of the story. Response A, on the other hand, doesn't even address the user's query about summarizing the book's plot.}\\
\bottomrule
\end{tabular}
}
\caption{The scoring guidelines and judgment for the ``Contextuality'' evaluation criterion.}
\label{tab:app_4}
\end{table*}

\begin{table*}[t]
\centering
\scalebox{0.85}{
\begin{tabular}{ p{2cm} | p{13cm} }
\toprule
\textbf{Data} & \textbf{Content} \\
\midrule
Criteria & \textbf{Conciseness}: \textit{Evaluates if the response is succinct yet comprehensive, providing a complete summary without unnecessary digression or redundancy.} \\
\midrule
Scoring Guidelines & \textbf{1. Not at all satisfactory}: \textit{The response doesn't summarize the plot of 'The Great Gatsby' or it includes wrong details. It rambles or digresses significantly from the main request, or is overly verbose without providing sufficient detail about the story.}

\textbf{2. Not very satisfactory}: \textit{The response gives a vague or partial plot of the book. There's some digression or redundancy, but some key elements of the story are mentioned. Some unnecessary details make the summary less concise.}

\textbf{3. Moderately satisfactory}: \textit{The response summarizes the main plot of 'The Great Gatsby', but it might miss some important details or contain some unnecessary information. There's a balance between conciseness and comprehensiveness.}

\textbf{4. Very satisfactory}: \textit{The response is fairly concise yet comprehensive. It manages to summarize the main plot of 'The Great Gatsby' without much digression. It might miss minor details or add a bit of extra information, but overall it maintains good balance between brevity and thoroughness.}

\textbf{5. Extremely Satisfactory}: \textit{The response perfectly summarizes 'The Great Gatsby'. It's concise, sticking to the important details only, yet comprehensive, covering all key aspects of the plot. It doesn’t include any unnecessary information or digress from the query, achieving a perfect balance of conciseness and comprehensiveness.}

\textbf{In Summary}:
\textit{This scoring guideline serves as a standard when evaluating responses, making sure they cater to the user query while adhering to the chosen evaluation criteria, i.e., conciseness and comprehensiveness in this case. Comparatively, a response scoring 1 or 2 on this scale would need significant improvement, while a response scoring 3 demonstrates average adherence. A score of 4 indicates that the response was strong, whereas a score of 5 should be reserved for responses that exemplify an ideal balance between conciseness and comprehensiveness.} \\

\midrule
Judgement &\textbf{Response A Score: 1}

\textbf{Explanation}: \textit{Response A doesn't provide a strong and concise summary of 'The Great Gatsby'. It does not adhere to user's request of summarizing the plot. Rather, it digresses into talking about the author, which is irrelevant. The story of the book itself is not explained at all, which is a clear deviation from the user query. Hence, falling into the first category of our scoring guideline.}

\textbf{Response B Score: 5}

\textbf{Explanation}: \textit{Response B manages to balance conciseness and comprehensiveness perfectly, as defined in category five of the scoring guidelines. It discusses the key elements of the plot of 'The Great Gatsby' succinctly. It talks about the protagonist, his love interest, as well as thematic elements of the book like false values of the rich and Gatsby's tragic end, without unnecessary information or digression.}

\textbf{Comparison}: \textit{The comparison of these two responses highlights the deficiencies in Response A. It doesn't address the user query at all, whereas Response B not only answers the query but does so in a succinct and comprehensive manner, striking a perfect balance between being brief yet detailed. Therefore, Response B should be considered as the ideal response.}\\

\bottomrule
\end{tabular}
}
\caption{The scoring guidelines and judgment for the ``Conciseness'' evaluation criterion.}
\label{tab:app_5}
\end{table*}

\begin{table*}[t]
\centering
\scalebox{0.85}{
\begin{tabular}{ l | p{14cm} }
\toprule
\textbf{Data} &  \textbf{Content} \\
\midrule
Response A Correction & ``The Great Gatsby'' by F. Scott Fitzgerald is set in the Roaring Twenties and revolves around the wealthy and mysterious Jay Gatsby, who is known for his lavish parties. The novel is narrated by Nick Carraway, Gatsby's neighbor, who recounts his observations about the lifestyle in Long Island's West Egg neighborhood. The plot thickens as it is revealed that Gatsby is ardently in love with Daisy Buchanan, who is married to the affluent and arrogant Tom Buchanan. Gatsby's elaborate attempts to win Daisy's love ultimately lead to tragic consequences. The story is a sharp critique of the American Dream, exploring themes of wealth, social status, love, and disillusionment. \\
\midrule
Response B Correction & The Great Gatsby is a riveting tale of Jay Gatsby, a mysterious millionaire, pining for his former lover, Daisy Buchanan. However, Daisy is now married to Tom Buchanan. Hoping to attract Daisy, Gatsby organizes extravagant parties, which emphasizes the extravagance and emptiness of the wealthy elite. The narrative delves into Gatsby's relentless pursuit of his romantic ideals, only to be met with tragic repercussions. More so, Gatsby's infatuation with Daisy also symbolizes his pursuit of the American Dream, a critical theme in the book. The story highlights how his quest for love intertwines with Daisy and Tom's, leading to a tragic climax where Gatsby is let down by those he trusted. The narrative comments on the hollowness of high society and explores themes such as wealth, love, and illusion, making it a poignant critique of the American Dream.\\
\bottomrule
\end{tabular}
}
\caption{Results of AI responses correction following the previous evaluation feedback.}
\label{tab:app_6}
\end{table*}

\end{document}